\documentclass[11pt,a4paper]{article}

\usepackage[hyphens]{url}
\usepackage[hyperref]{acl2017}
\usepackage{breakurl}
\usepackage{times}
\usepackage{latexsym}
\usepackage{color}
\usepackage{graphicx}
\usepackage{multirow}
\usepackage{tikz}

\aclfinalcopy 

\newcommand{\mr}[2]{\multirow{#1}{*}{#2}}

\title{Subregular Complexity and Deep Learning}

\author{
Enes Avcu \\
Department of \\
Linguistics and\\ 
Cognitive Science\\
University of Delaware \\
{\footnotesize\tt enesavc@udel.edu} \And
Chihiro Shibata \\
School of \\ 
Computer Science\\
Tokyo University of\\ 
Technology\\
{\footnotesize\tt shibatachh@stf.teu.ac.jp} \And
Jeffrey Heinz \\
Department of Linguistics\\
Institute of Advanced\\
Computational Science\\
Stony Brook University \\
{\footnotesize\tt jeffrey.heinz@stonybrook.edu} }

\date{}

\usepackage{amssymb,booktabs}

\providecommand{\leftend}{\ensuremath{\rtimes}} 
\providecommand{\rightend}{\ensuremath{\ltimes}}
\providecommand{\LE}{\leftend}
\providecommand{\RE}{\rightend}
\providecommand{\subseq}{\ensuremath{\mathtt{subseq}}}
\providecommand{\factor}{\ensuremath{\mathtt{factor}}}
\def\Nat{\mathbb{N}}

\begin{document}
\maketitle
\begin{abstract}
This paper argues that the judicial use of formal language theory and grammatical inference are invaluable tools in understanding how deep neural networks can and cannot represent and learn long-term dependencies in temporal sequences. 

Learning experiments were conducted with two types of Recurrent Neural Networks (RNNs) on six formal languages drawn from the Strictly Local (SL) and Strictly Piecewise (SP) classes. 
The networks were Simple RNNs (s-RNNs) and Long Short-Term Memory RNNs (LSTMs) of varying sizes. 
The SL and SP classes are among the simplest in a mathematically well-understood hierarchy of subregular classes. 
They encode local and long-term dependencies, respectively. 
The grammatical inference algorithm Regular Positive and Negative Inference (RPNI) provided a baseline.  

According to earlier research, the LSTM architecture should be capable of learning long-term dependencies and should outperform s-RNNs. 
The results of these experiments challenge this narrative. 
First, the LSTMs' performance was generally worse in the SP experiments than in the SL ones.
Second, the s-RNNs out-performed the LSTMs on the most complex SP experiment and performed comparably to them on the others.
\end{abstract}

\maketitle

\section{Investigating Deep Learning}

This paper argues that formal language theory and grammatical inference can provide a systematic way to better understand the kinds of patterns deep learning networks \cite{GoodfellowBengioCourville2016} are able to learn. 
The main ideas are illustrated with experiments testing how well two types of Recurrent Neural Networks (RNNs) can learn different kinds of simple, subregular formal languages with a grammatical inference algorithm serving as a baseline.

Using formal languages to investigate the learning capabilities of neural networks is not without precedent. 
Much earlier  research also used formal languages to probe the learning capabilities of neural networks; \citet[sec.~5.13]{Schmidhuber2015} provides a review.
Section~\ref{sec:mot} highlights some of this work and makes clear our own contribution.

Long-term dependencies in temporal sequences have a distinguished history in the development of neural network learning models and in generative linguistics. 
\citet{BENGIO1994} define long-term dependencies this way: ``A task displays long-term dependencies if prediction of the desired output at time $t$ depends on input presented at an earlier time $\tau\ll t$.'' 
Many examples of such long-term dependencies abound in nature and engineering. For example, generative linguists, beginning with \citet{chomsky56,chomsky57}, have studied the grammatical basis of long-term dependencies in natural languages and have raised the question of how such dependencies are learned \citep{chomsky65}.

We test simple RNNs (s-RNNs) \citep{ELMAN1990179} and Long Short-Term Memory RNNs (LSTMs) \citep{lstm1997} on simple regular languages which encode local and long-term dependencies. 
Readers are referred to \citet{GoodfellowBengioCourville2016} and \citet{Goldberg2017} for details of these two types of networks.

A common narrative in the deep learning literature is that LSTMs are a solution to learning long-term dependencies, which are problematic for s-RNNs. 
For example, Schmidhuber's (\citeyear{Schmidhuber2015}) review, which received the first Best Paper Award ever issued by the journal \emph{Neural Networks}, explains that ``Typical deep NNs suffer from the now famous problem of vanishing or exploding gradients.''  
He calls this ``the fundamental deep learning problem of gradient descent.'' 
It is these vanishing or exploding gradients that prevent neural networks like s-RNNs from learning long-term dependencies.  
Schmidhuber explains how much subsequent research was dedicated to overcoming this problem and writes ``LSTM-like networks \ldots alleviate the problem through a special architecture unaffected by it.''

Similarly, writing in \emph{Nature}, \citet[p.~442]{deep} say ``Although [RNNs] main purpose is to learn long-term dependencies, theoretical and empirical evidence shows that it is difficult to learn to store information for very long.'' 
They go on to write ``LSTM networks have subsequently proved to be more effective than conventional RNNs'' because LSTMs ``use special hidden units, the natural behaviour of which is to remember inputs for a long time.'' 

Therefore, we were particularly interested in understanding how well LSTMs can learn long-term dependencies within temporal sequences.
We developed training and test data sets for formal languages drawn from the Strictly Local (SL) and Strictly Piecewise (SP) classes of formal languages.
As will be explained in more detail in \S\ref{sec:subreg}, SL and SP languages are simple regular languages which only encode local and certain types of long-term dependencies, respectively.  

These formal languages are drawn from well-understood subclasses of the regular languages which form a complexity hierarchy \citep{McNaughtonPapert1971,Rogers-HeinzEtAl-2010-LPTSS,RogersPullum2011}.
These hierarchies measure the complexity of formal languages not in terms of automata-theoretic measures, such as the size of the minimal deterministic automaton, but instead on a model-theoretic basis \citep{Enderton2001,Rogers-HeinzEtAl-2013-CSC}. 
In other words, the complexity of a formal language is determined by the kind of logic and model-theoretic representation needed to specify it \citep{RogersPullum2011}. 
As \citet{Rogers-HeinzEtAl-2013-CSC} explain, these classes also have a cognitive interpretation.

In the experiments, there were six target languages to learn: three SL and three SP. 
For each language, three training sets were prepared, and for each training set two test sets were prepared, for a total of 36 test sets. 
The training and test data was also controlled for word length so we could assess the networks' ability to generalize to strings longer than the ones in the training sample.
The LSTMs and s-RNNs were trained on both positive and negative examples.
We conducted several experiments, systematically varying the vector sizes in the networks.
These experimental details are explained in \S\ref{sec:exp}. 

The results, presented in \S\ref{sec:result}, are unexpected given the narrative outlined above. 
The narrative would suggest that the s-RNNs and LSTMs may perform comparably on the SL experiments, but that s-RNN performance would be worse than LSTM performance on the SP experiments due to the presence of long-term dependencies. 
Furthermore, since the LSTMs are ``unaffected'' by the ``fundamental deep learning problem,'' we may expect that the LSTM performance on the SP experiments to be comparable to the ones on the SL experiments.

Neither of these expectations were borne out. 
While the RNNs performed above chance in all of our experiments, they struggled learning the two most complex SP languages as compared to the matched SL languages.
Furthermore, the s-RNNs performed comparably to the LSTMs in many of the SP experiments, and in fact out-performed the LSTMs on the most complex SP learning task. 
Also, both LSTMs and s-RNNs did relatively poorly on the simplest SL experiment.

When learning fails, it is natural to ask whether the training data was sufficiently rich for it to be reasonable for correct inference to take place. 
For this reasons, we also ran the grammatical inference algorithm  Regular Positive and Negative Inference (RPNI) \citep{OncinaGarcia1992-RPNI} on the test sets and examined its output. 
RPNI provably infers any regular language, provided the training data is sufficient. 
Readers are referred to \citet{Higuera2010} for details on RPNI. 
When RPNI is successful, it means there is enough information in the training sample for correct inference to occur, at least for learning algorithms which only consider regular languages as targets. 
RPNI's results suggest that training data was sufficient in almost all of the SP experiments, but only in one-third of the SL experiments. This one-third includes the simple SL experiments where the RNNs struggled.
The analysis with RPNI makes it harder to explain away the poor performance of the RNNs on the grounds that the data was insufficient.
Section~\ref{sec:disc} discusses this, and other aspects of the results in more detail.

Our conclusion is that the there is much more to be learned about the how RNNs represent and learn long-term dependencies in sequences. 
We believe that understanding how RNNs generalize from their training data will follow from connecting the behavior of RNNs to classes of formal languages like the ones here.
 
\section{Motivation and background}
\label{sec:mot}

In the 1990s, many studies aimed to learn formal languages with neural networks. 
When the aim was to predict the next symbol of a string drawn from a regular language, first-order RNNs were used \cite{Casey1996,SmithA.W.1989}. 
The target languages here were based on the Reber grammar \cite{REBER1967}. 
When the aim was to decide whether a string is grammatical, second-order RNNs were used \cite{Pollack1991,Watrous1992,Giles1992}. 
Here the target languages were the regular languages studied by Tomita \cite{Tomita1982}. 
Later research targeted nonregular languages \cite{Schmidhuber2002,Chalup2003955}. One striking result established that LSTMs can learn some context-sensitive formal languages exhibiting long-distance dependencies with uncanny precision \cite{PérezOrtiz2003241}. 

The reasons for making formal languages the targets of learning are as valid today as they were decades ago. 
First, the grammars generating the formal languages are known. 
Therefore training and test data can be generated as desired. 
Thus, the scientist can run controlled experiments to see whether particular generalizations are reliably acquired under particular training regimens. 

Importantly, the relative complexity of different formal languages may provide additional insight.
If it is found that formal languages of one type are more readily learned than formal languages of another type in some set of experiments then the difference between these classes may be said to meaningfully capture some property unavailable to the RNNs in the experiments.
Subsequent work may lead to proofs and theorems about which properties of RNNs lead to the reliable inference of formal languages from certain classes and which do not. 
It may also lead to new network architectures which overcome identified hurdles. 

There are two important differences between the present paper and past research, beyond the development in neural networks. 
First, the regular languages chosen here are known to have certain properties. 
The Reber grammars and Tomita languages were not understood in terms of their abstract properties or pattern complexity. 
While it was recognized some encoded long-term dependencies and some did not, there was little recognition of the computational nature of these formal languages beyond that. 
In contrast, the formal languages in this paper are much better understood. 
While \emph{subregular} distinctions had already been studied by the time of that research \cite{McNaughtonPapert1971}, it went unrecognized how that branch of computer science could inform neural network learning. 

The second difference is the advances in grammatical inference over the past few decades. The development of RPNI \citep{OncinaGarcia1992-RPNI} essentially solved the problem of efficiently learning regular languages from positive and negative data. Other results addressed the learning of subregular classes from positive data only \citep{GarciaEtAl1990,Garca2004LearningKA,Heinz-2010-SEL,Heinz-KasprzikEtAl-2012-LLHS}. Like the work on subregular complexity, what these analytical approaches to learning formal languages offered neural network researchers went unrecognized.

\section{Subregular Complexity}
\label{sec:subreg}

Figure~\ref{fig:subreg} shows proper inclusion relationships of well-studied classes of subregular languages. The Strictly Local (SL), Locally Testable (LT), and Non-Counting (NC) classes were studied by \cite{McNaughtonPapert1971}. 
The Locally Threshold Testable (LTT) class was introduced and studied by \cite{Thomas1982}. 
The Piecewise Testable (PT) class was introduced and studied by \cite{Simon1975}. 
The Strictly Piecewise (SP) class was studied by \cite{Rogers-HeinzEtAl-2010-LPTSS}.  
As many authors discuss, these classes are natural because they have multiple characterizations in terms of logic, automata, regular expressions, and abstract algebra. Cognitive interpretations of these classes also exist \cite{RogersPullum2011,Rogers-HeinzEtAl-2013-CSC}.

From the perspective of natural language processing, SL is the formal language-theoretic basis of n-gram models \cite{JM2008} and SP models aspects of phonology \cite{Heinz-2010-LLP}.
\begin{figure*}
\centering
\begin{tikzpicture}[description/.style={fill=white,inner sep=2pt},scale=0.5]
  \node (REG)  at (0,7) {Regular};
  \node (SF)   at (4,5) {Non-Counting};
  \node (LTT)  at (-4,4) {Locally Threshold Testable};
  \node (LT)   at (-4,2) {Locally Testable};
  \node (PT)   at (4,2) {Piecewise Testable};

  \node (SL)   at (-4,-1) {Strictly Local};
  \node (SP)   at (4,-1) {Strictly Piecewise};

  \node (succ) at (-4,-3) {\textbf{Successor}};
  \node (prec) at (4,-3) {\textbf{Precedence}};

  \node (mso1)        at (11,8)   {\textbf{Monadic}};
  \node (ms02)        at (11,7) {\textbf{Second Order}}; 
  \node (fo1)         at (11,5) {\textbf{First}};
  \node (fo1)         at (11,4) {\textbf{Order}}; 
  \node (prop)       at (11,2)   {\textbf{Propositional}};
  \node (cnl1)        at (11,0)   {\textbf{Conjunctions}};
  \node (cnl2)        at (11,-1)   {\textbf{of Negative}};
  \node (cnl3)        at (11,-2)   {\textbf{Literals}};

  \draw (REG) -- (SF) -- (LTT) -- (LT) -- (SL);
  \draw (REG) -- (LTT);
  \draw (SF) -- (PT) -- (SP);
  \draw (SF) -- (PT);

  \draw[blue, thick] (-9,1)  -- (8,1);
  \draw[blue, thick] (-9,3)  -- (8,3);
  \draw[blue, thick] (-9,6)  -- (8,6);

  \draw[blue, thick] (8,-2) -- (-9,-2);
  \draw[blue, thick, ->] (-9,-2) -- (-9,8);

  \node[rotate=90] (complexity)       at (-10,2)   {\textit{Complexity}};

 \node[rotate=90] (logic)       at (-11,2)   {\textit{Logical}};
 
 \node (order)       at (0,-5)   {\textit{Representation of Order}};
\end{tikzpicture}
\caption{Subregular language classes  with inclusion shown from the top down.}
\label{fig:subreg}
\end{figure*}

Next we define these classes, focusing on the SL and SP classes since languages belonging to them form the learning targets in the experiments described in \S\ref{sec:exp}.

\subsection{Mathematical Notation}

Let $\Sigma$ denote a finite set of symbols, the alphabet, and
$\Sigma^*$ the set of elements of the free monoid of $\Sigma$ under
concatenation.  We refer to these elements both as \emph{strings} and
as \emph{words}.   
The $i$th symbol in word $w$ is
denoted $w_i$. Left and right word boundary
markers (\LE~and \RE, respectively) are symbols not in $\Sigma$. A stringset (also called formal language) is a subset of $\Sigma^*$.

If $u$ and $v$ are strings, $uv$ denotes their concatenation. 
Similarly, if $S_1,S_2$ are stringsets then $S_1S_2$ denotes their concatenation and is equal to $\{uv\mid u\in S_1, v\in S_2\}$. 

For all $u,v,w,x\in\Sigma^*$, if $x=uwv$ then then $w$ is a \emph{substring} of $x$. 
If $x\in\Sigma^*w_1\Sigma^*w_2\Sigma^*\ldots w_n\Sigma^*$ then $w$ is a \emph{subsequence} of $x$. 
A substring (subsequence) of length $k$ is called a $k$-factor ($k$-subsequence).
Let $\factor_k(w)$ denote the set of substrings  of $w$ of length $k$. 
Let $\subseq_k(w)$ denote the set of subsequences of $w$ up to length $k$. The domains of these functions extend to languages in the normal way.

\subsection{Strictly Local Stringsets} 

A stringset $L$ is Strictly $k$-Local (SL$_k$) iff whenever there is a string x of length $k-1$ and strings $u_1, v_1, u_2, v_2\in \Sigma^*$, such
that $u_1 xv_1, u_2 xv_2\in L$ then $u_1 xv_2 \in L$. We say $L$ is \emph{closed under suffix substitution}. $L$ is SL if $L\in$ SL$_k$ for some $k$ \cite{RogersPullum2011}.

As discussed in \cite{McNaughtonPapert1971,RogersPullum2011}, SL$_k$ languages can also be characterized by a finite set of  $k$-factors as follows. 
Observe first that for each $k$, $\factor_k(\{\LE\}\Sigma^*\{\RE\})$ is finite.
Let a SL$_k$ grammar be a set of $k$-factors $G\subseteq\factor_k(\{\LE\}\Sigma^*\{\RE\})$. 
The language of G is the stringset $L(G) = \{ w\mid \factor_k(\LE w\RE)\subseteq G\}$. 
Thus, the grammar $G$ is the set of \emph{permissible} $k$-factors. 
Any $k$-factor $w$ in $\factor_k(\{\LE\}\Sigma^*\{\RE\})$ which is not in $G$ is thus \emph{forbidden} and consequently all strings containing $w$ as a substring are not in $L(G)$.
Consequently, SL stringsets can also be defined with grammars that only contain finitely many forbidden  $k$-factors as we do in \S\ref{sec:exp}. From a logical perspective, SL stringsets can thus be expressed as the conjunction of negative literals where literals correspond to a model-theoretic representation of strings where the order of the elements is given by a successor relation \citep{RogersPullum2011}.


\subsection{Strictly Piecewise Stringsets}

A stringset $L$ is Strictly $k$-Piecewise (SP$_k$) iff $\subseq_k(w)\subseteq\subseq_k(L)$ implies $w\in L$. 
$L$ is SP if there is a $k$ such that it belong to SP$_k$; equivalently, $L$ belongs to SP iff $L$ is closed under subsequence \cite{Rogers-HeinzEtAl-2010-LPTSS}.
SP$_k$ stringsets can also be defined with a finite set of $k$-subsequences \cite{Rogers-HeinzEtAl-2010-LPTSS}. 
In fact the parallel to SL$_k$ is near perfect. 

Observe the set $\subseq_k(\Sigma^*)$ is finite.
Let a SP$_k$ grammar be a set of $k$-subsequences $G\subseteq\subseq_k(\Sigma^*)$.
The language of G is the stringset $L(G) = \{ w\mid \subseq_k(w)\subseteq G\}$. 
The grammar $G$ is the set of \emph{permissible} $k$-subsequences. 
Any $k$-subsequence $w$ in $\subseq_k(\Sigma^*)$ which is not in $G$ is thus \emph{forbidden} so strings containing $w$ as a subsequence are not in $L(G)$.
Since for each $k$, $\subseq_k(\Sigma^*)$ is finite, SP stringsets can also be defined with grammars containing forbidden $k$-subsequences as in ~\S\ref{sec:exp}.
From a logical perspective, SP stringsets can thus be expressed as the conjunction of negative literals where literals correspond to a model-theoretic representation of strings where the order of the elements is given by the precedence relation \citep{Rogers-HeinzEtAl-2013-CSC}.

\subsection{Locally and Piecewise Testable Classes}

A stringset $L$ is Locally $k$-Testable (LT$_k$) iff for all $w,v\in\Sigma^*$, it is the case that if $\factor_k(\LE w\RE)=\factor_k(\LE v\RE)$ then either $w,v\in L$ or $w,v\not\in L$. In other words, membership of a string $w$ in any LT$_k$ stringset is determined solely by the set of $k$-factors in $w$. Similarly, a stringset $L$ is Piecewise $k$-Testable (PT$_k$) iff for all $w,v\in\Sigma^*$, it is the case that if $\subseq_k(\LE w\RE)=\subseq_k(\LE v\RE)$ then either $w,v\in L$ or $w,v\not\in L$. Here, membership of a string $w$ in any PT$_k$ stringset is determined solely by the set of $k$-subsequences in $w$. Stringsets are LT (PT) if there is some $k$ such that they are LT$_k$ (PT$_k$), respectively.

It can be proven that LT (PT) languages are those equivalent to a Boolean combination of finitely many SL (SP) formal languages. LT (PT) are also parameterized by a $k$ value, which corresponds to the largest $k$-value among the SL languages they are a Boolean combination of.

From a logical perspective, the LT and PT classes can be defined with propositional statements over relational structures representing sequences where the order relation is given as successor and precedence, respectively,  as shown in Figure~\ref{fig:subreg} \citep{Rogers-HeinzEtAl-2013-CSC}.

\subsection{Locally Threshold Testable, NonCounting, and Regular Classes}

A stringset $L$ is Locally Threshold Testable iff there are two numbers $k$ and $t$ such that for all strings $u,v\in\Sigma^*$ and $k$-factors $x\in\factor_k(\{\LE\}\Sigma^*\{\RE\})$  if the number of times $x$ occurs in $u$ is the same as the number of times $x$ occurs in $v$ whenever this number is less than $t$ or occurs at least $t$ times in both $u$ and $v$ then either $u,v\in L$ or $u,v\not\in L$. 
In other words,  membership of a string $w$ in any LTT$_{t,k}$ stringset is determined solely by the number of occurences each $k$ factor occurs in $w$, counting them only up to some threshold $t$.
The LT$_k$ class equals LTT$_{1,k}$. 

A stringset $L$ is NonCounting iff there is a $k$ such that for all $w,u,v\in\Sigma^*$, if $wuv\in L$ then $wu^{k+1}v\in L$. 
\citet{McNaughtonPapert1971} prove languages in the NonCounting class are exactly those definable with star-free generalized regular expressions and exactly those obtained by closing LT stringsets under concatenation. 

Languages in the LTT and NC classes can be defined with first-order statements over relational structures representing sequences where the order relation is given as successor and precedence, respectively \citep{Thomas1982,McNaughtonPapert1971,Rogers-HeinzEtAl-2013-CSC}. 
LTT is thus properly included in NC because successor is first-order definable with precedence but precedence is not first-order definable with successor.

Informally, a stringset $L$ is regular if the resources required to decide whether a string $u$ belongs to $L$ is independent of the length of $u$. 
They can be defined as those formal languages recognizable by finite-state acceptors. 
\citet{Buchi1960} showed these are exactly the stringsets definable with weak monadic second-order logic with the order relation given as successor (or precedence, since the precedence relation is MSO-definable from successor and vice versa). 
Stringsets that are regular but not NonCounting typically  count modulo some $n$. 
For example, the stringset which contains all and only strings with an even number of \texttt{a}s is not NonCounting, but regular. 

\subsection{Further Comment}

SL, SP, LT, PT, and LTT classes form infinite hierarchies of language classes based on $k$ \cite{RogersPullum2011,Rogers-HeinzEtAl-2010-LPTSS}. 
In particular for all $k\in\Nat$, SL$_k\subsetneq$ SL$_{k+1}$,  SP$_k\subsetneq$ SP$_{k+1}$, LT$_k\subsetneq$ LT$_{k+1}$, PT$_k\subsetneq$ PT$_{k+1}$ and LTT$_{t,k}\subsetneq$ LTT$_{t,k+1}$. It is also true that LTT$_{t,k}\subsetneq$ LTT$_{t+1,k}$. 
Consequently, for any SL (SP/LT/PT/LTT) stringset, the smallest $k$ value for which it is SL$_k$ (SP$_k$/LT$_k$/PT$_k$/LTT$_{t,k}$) is another measure of its complexity.

We analyzed the \citet{REBER1967} and \citet{Tomita1982} languages and concluded the following. 
The Reber grammar is SL$_3$ and the embedded Reber grammar is LT$_3$. 
Tomita languages 1, 2, 3, 4, 5, 6, and 7 are SL$_1$, SL$_2$, regular, SL$_3$, regular, regular, and SP$_4$, respectively.

The subregular hierarchies provide a much more fine-grained meaning to the term \emph{long-term dependency}.
For example, SL dependencies are effectively bounded by a window of size $k$, but none of the other classes are limited in this way. 
Consequently the  distinctions between them distinguish different \emph{kinds} of long-term dependencies. 
This is why we are optimistic that much these subregular properties can meaningfully inform how RNNs represent and learn long-term dependencies.

\section{The Experiments}
\label{sec:exp}
Here we describe the target languages, the training data, the test sets, the neural network architectures and RPNI.

We implemented the RNNs with  \texttt{Chainer} (\url{http://chainer.org}). 
RPNI was implemented using Matlab and the \texttt{gi-toolbox} 
\url{https://code.google.com/archive/p/gitoolbox}
\citep{Akram2010}. The files we used with \texttt{Chainer}, \texttt{gi-toolbox}, and the ones we used to prepare the training and test sets are available online at \url{https://github.com/enesavc/subreg_deeplearning}.

\subsection{Target Languages}

In this study, six formal target languages were defined in order for training and testing purposes.  
In each case, we let $\Sigma=\{a,b,c,d\}$.

Table~\ref{tab:lgs} defines the six formal languages we used in this study. Each of the SL languages was defined with four banned substrings and each of the SP languages was defined with one banned subsequence.  We refer to these six languages by the class they belong to: SL2, SL4, SL8, SP2, SP4, and SP8.
\begin{table*}[t]
\centering
\caption{The six target stringsets with $\Sigma=\{a,b,c,d\}$.}
\label{tab:lgs}
\begin{tabular}{llc}
\toprule
Language Class & Forbidden $k$-factors in target stringsets& Minimal DFA size\\
\midrule
SL2 & \texttt{\LE b, aa, bb, a\RE}&3\\
SL4 & \texttt{\LE bbb, aaaa, bbbb, aaa\RE}&7\\
SL8 & \texttt{\LE bbbbbbb, aaaaaaaa, bbbbbbbb, aaaaaaa\RE}&15\\
\midrule
Language Class & Forbidden $k$-subsequences in target stringsets\\
\midrule
SP2 & \texttt{ab}&2\\
SP4 & \texttt{abba}&4\\
SP8 & \texttt{abbaabba}&8\\\bottomrule
\end{tabular}
\end{table*}

These grammars were implemented as finite-state machines using \texttt{foma}, a publicy available, open-source platform \cite{hulden2009-foma}. 
The number of states in the minimal deterministic automaton recognizing these six languages are shown in Table~\ref{tab:lgs}.

\subsection{Training data}

Training data was generated with \texttt{foma}. 
For each language L we generated three training data sets, which we call 1k, 10k, and 100k because they contained 1,000, 10,000, and 100,000 words, respectively. 
Half of the words in each training set were positive examples (so they belonged to L) and half were negative examples (so they did not belong to L). 
Training words were between length 1 and 25. 
For the positive examples there were 20, 200, and 2,000 words of each length. 
For the negative examples, we wanted to provide 20, 200, and 2,000 words of each length, respectively. 
However, as the $k$ value increases, there is no negative data for shorter words since all shorter words belong to L. 
In this case, we generated 20$\times$k, 200$\times$k, and 2,000$\times$k of words of length $k$, and also 20, 200, and 2,000 words for the lengths between $k+1$ and 25.
Training words were generated randomly using \texttt{foma}, so training sets contained duplicates. 
\subsection{Test sets}

For each language L and each training regimen T for L, we developed two test sets, which we call Test1 and Test2. 
Test1 and Test2 contain 1,000, 10,000, or 100,000 words depending on whether T is 1k, 10k, or 100k, respectively. 
Half of the test words belong to L and half do not.
Test1 and Test2 only contain novel words. 
Novel positive words belong to L but do not belong to the positive examples in T.
Novel negative words neither belong to L nor to the negative examples in T. 

The difference between the two test sets has to do with word length. 
Test1 words are no longer than 25. 
Test2 words are of length between 26 and 50. 
Thus while both sets test for generalization, Test2 importantly checks to what extent the generalizations are independent of word length.

For Test2, we used \texttt{foma} to randomly generate 20, 200, and 2,000 positive words and negative words for each length between 26 and 50. 

For Test1, we used \texttt{foma} to randomly generate words whose length was less than 26. 
Again, we wanted to generate 20, 200, and 2,000 positive words and negative words for each length. However, there may not be positive or negative words for some of the shorter lengths. 
This is because either they do not exist for the target language L, or because they do exist for L but they also exist in the training data in which case they would not be novel. 
For positive (negative) words, we did the following. 
If there was at least one word of a particular length, we randomly generated 20, 200, or 2,000 words of that length depending on T. 
We also generated extra words of length 25 (specifically 200, 2,000, or 20,000). 
We concatenated these together ordered by length into one file and then selected the first 500, 5,000, or 50,000 words to be the positive (negative) words in Test1.
In other words, we effectively padded the Test1 with words of length 25 when there were no words of shorter lengths.

The order of the words in the test sets was randomized.

\subsection{RNN Architectures}

For the LSTMs and s-RNNs, we constructed simple networks to test the capability of the networks themselves. 
First we describe properties of the architectures shared by both RNN types (the LSTMs and s-RNNs) and then we discuss specific aspects of the LSTMs.

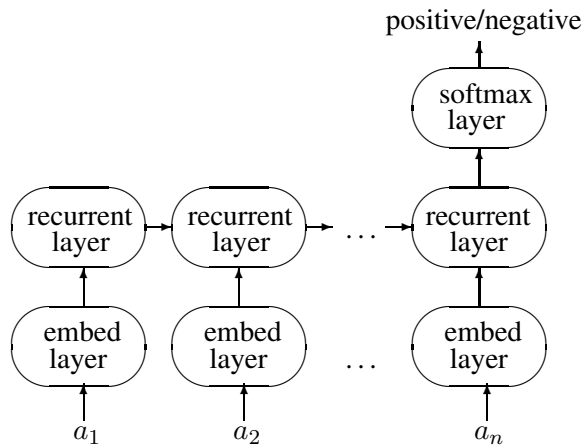
\begin{figure}
\centering
\begin{picture}(180,160)
\put(0,80){\oval(50,30)}
\put(60,80){\oval(50,30)}
\put(150,80){\oval(50,30)}
\put(-19,81){recurrent}
\put(41,81){recurrent}
\put(130,81){recurrent}
\put(-10,71){layer}
\put(50,71){layer}
\put(139,71){layer}

\put(150,95){\vector(0,1){15}}
\put(150,125){\oval(50,30)}
\put(135,127){softmax}
\put(138,117){layer}
\put(150,140){\vector(0,1){10}}
\put(115,155){positive/negative}

\put(25,80){\vector(1,0){10}}
\put(85,80){\vector(1,0){10}}
\put(115,80){\vector(1,0){10}}
\put(100,76){\ldots}

\put(0,35){\oval(50,30)}
\put(60,35){\oval(50,30)}
\put(150,35){\oval(50,30)}
\put(-13,37){embed}
\put(47,37){embed}
\put(137,37){embed}
\put(-11,27){layer}
\put(49,27){layer}
\put(139,27){layer}
\put(100,27){\ldots}
\put(2,50){\vector(0,1){15}}
\put(60,50){\vector(0,1){15}}
\put(150,50){\vector(0,1){15}}

\put(-2,0){$a_1$}
\put(58,0){$a_2$}
\put(148,0){$a_n$}
\put(2,7){\vector(0,1){14}}
\put(62,7){\vector(0,1){14}}
\put(153,7){\vector(0,1){14}}

\end{picture}
\caption{An unrolled RNN for predicting membership.}
\label{fig:lstm1}
\end{figure}

For each input string, the RNNs can be represented as a connected graph as shown in Figure~\ref{fig:lstm1}. The output of the network is the probability of the input word belonging to the target language.

The output of the embed layer in Figure~\ref{fig:lstm1} is known as the distributed representation of the symbol, 
which is equivalent to a linear layer (with no bias) that maps the one-hot vector to a real-valued vector. 
The outputs of the recurrent layer corresponding to each symbol are ignored except for the last one. 
The output corresponding to the last symbol is mapped to a two dimensional vector through the softmax layer, whose elements represent the positive and negative probabilities. 

Each value of the weights in the embed layer were independently initialized according to the normal distribution.

For all the RNNs in this study the vector sizes of the recurrent layer and the embed layer were made identical. 
For each type of RNN, we examined three networks of varying sizes, which we call v10, v30, and v100 because the vectors at each layer are of size 10, 30, and 100, respectively.

For the LSTMs, the standard architecture with forget gates and without peepholes was used among the various possible modifications~\cite{GreffSKSS15}.
The recurent layer of the LSTMs includes four fully-connected layers  called the input gate, the block input, the forget gate, and the output gate.
The weights of those layers are initialized in the same manner as the embed layer except for the weights of the forget gate, which were initialized according to the normal distribution with mean 1 and variance 1.




\subsection{RNN Training}

When training the RNNs, the training data was divided into batches, and the gradient is calculated for each batch.
The strings in each batch are usually chosen uniformly and randomly to avoid biased gradients.
However, calculations become inefficient if strings are chosen completely randomly due to large differences in the lengths of the strings. 
Thus, we sort all strings first in order of their lengths, re-order partially, block them and choose those blocks randomly. The RNNs processed the training data 100 times or `epochs.'

Other training parameters are as follows.
The batch size is $128$. 
The L2 norm of the gradient is clipped with $1.0$.
The lengths of strings in each batch are aligned through padding an additional symbol whose embedding is fixed to the zero vector. The optimization algorithm called Adam~\cite{KingmaB14} is applied.

\subsection{RPNI}

RPNI takes as input a training set of positive and negative examples and outputs a deterministic finite-state acceptor (DFA), which accepts and rejects strings. 

RPNI itself is deterministic. 
In other words, if RPNI is run on the same training data multiple times, the variance in its performance on the test sets will be zero. 
This is because for a given training set, it will always output the same DFA.

RPNI is a state-merging algorithm. 
It first builds a DFA which explicitly accepts and rejects only the positive and negative examples in training. 
This DFA is known as a prefix-tree because its states are in one-to-one correspondence with the prefixes of the strings in the training set.
It then conducts a breadth-first traversal of the prefix tree, attempting to successively merge pairs of states. 
Two states are merged provided the resulting DFA is consistent with training set.
Readers are referred to \citet{Higuera2010} for more details on RPNI and to  \citet{Heinz-delaHiguera-vanZaanen-GICL} for more general discussion of state-merging algorithms. 

\citet{OncinaGarcia1992-RPNI} proved that for each regular language $L$ there is a finite set $S^+$ of positive examples and a finite set $S^-$ of negative examples such that RPNI, when given any training set including $S^+$ and $S^-$ as input, will output a minimal DFA $A$ recognizing $L$. 
Furthermore, RPNI runs in cubic time with respect to the size of $S^+\cup S^-$. Additionally, the size of $S^+\cup S^-$ is also bounded by a polynomial with respect to the size of $A$. Readers are referred to \citet{Higuera1997} and \citet{Eyraud-HeinzEtAL-2016-EILLP} for more details on analyses of efficiency in grammatical inference.

\section{Results}
\label{sec:result}
 
After each epoch, the RNNs were tested on the test sets. 
Inspection of the accuracy trajectories indicate that accuracy had stabilized in all cases well before the 100th epoch. 
Therefore, accuracy results are reported after the 100th epoch of training.
\begin{table*}[t]
\centering
\caption{Accuracy on Target SL Stringsets after 100 Epochs}
{\footnotesize
\label{tab:resultsSL}
\setlength\tabcolsep{4.5pt}
\begin{tabular}{ccc|ccc|ccc|c}
\toprule
\multicolumn{2}{c}{\mr{2}{Training}} & \mr{2}{Test} & 
\multicolumn{3}{c|}{LSTM}            & \multicolumn{3}{c|}{s-RNN}
                                     & \mr{2}{RPNI} \\[0.5ex] 

            &              &   & 10          & 30          & 100              & 10            & 30            & 100              &          \\\midrule
\mr{6}{SL2} & \mr{2}{1k}   & 1 & 0.772 (0.09) & 0.717 (0.08) & 0.711 (0.02)   & 0.766 (0.11)  & 0.761 (0.11)  & 0.762 (0.10)     & \bf 0.855     \\ 
            &              & 2 & 0.758 (0.09) & 0.696 (0.10) & 0.685 (0.02)   & 0.757 (0.15)  & 0.784 (0.17)  & 0.768 (0.15)     & \bf 0.844     \\\cmidrule{3-10} 
            & \mr{2}{10k}  & 1 & 0.773 (0.17) & 0.616 (0.01) & 0.666 (0.01)   & 0.682 (0.15)  & 0.660 (0.11)  & 0.649 (0.11)     & \bf 1.000 \\ 
            &              & 2 & 0.772 (0.19) & 0.602 (0.01) & 0.650 (0.01)   & 0.675 (0.16)  & 0.650 (0.12)  & 0.639 (0.12)     & \bf 1.000 \\\cmidrule{3-10}  
            & \mr{2}{100k} & 1 & 0.684 (0.15) & 0.615 (0.03) & 0.644 (0.01)   & 0.700 (0.14)  & 0.723 (0.16)  & 0.620 (0.01)     & \bf 1.000 \\ 
            &              & 2 & 0.669 (0.16) & 0.596 (0.02) & 0.624 (0.01)   & 0.689 (0.16)  & 0.718 (0.18)  & 0.601 (0.01)     & \bf 1.000 \\\midrule
\mr{6}{SL4} & \mr{2}{1k}   & 1 & 0.902 (0.01) & 0.907 (0.07) & 0.884 (0.06)   & 0.913 (0.01)  & 0.956 (0.01)  & \bf 0.968 (0.01) & 0.918     \\ 
            &              & 2 & 0.836 (0.01) & 0.890 (0.04) & 0.901 (0.02)   & 0.844 (0.01)  & 0.896 (0.01)  & \bf 0.911 (0.01) & 0.813     \\\cmidrule{3-10}  
            & \mr{2}{10k}  & 1 & 0.840 (0.15) & 0.856 (0.12) & 0.942 (0.08)   & 0.934 (0.12)  & 0.982 (0.00) & 0.977 (0.01)      & \bf 0.995     \\                
            &              & 2 & 0.836 (0.16) & 0.852 (0.13) & 0.938 (0.08)   & 0.938 (0.12)  & \bf 0.993 (0.00) & 0.991 (0.00)  & 0.978     \\\cmidrule{3-10}  
            & \mr{2}{100k} & 1 & 0.975 (0.05) & 0.917 (0.12) & 0.898 (0.10)   & 0.905 (0.16)  & 0.989 (0.00) & 0.986 (0.00)      & \bf 1.000 \\ 
            &              & 2 & 0.981 (0.04) & 0.923 (0.12) & 0.903 (0.10)   & 0.916 (0.16)  & 0.995 (0.00) & 0.994 (0.00)      & \bf 1.000 \\\midrule
\mr{6}{SL8} & \mr{2}{1k}   & 1 & 0.981 (0.02) & 0.976 (0.04) & 0.995 (0.00)   & 0.989 (0.01)  & \bf 0.999 (0.00) & \bf 0.999 (0.00) & 0.991     \\ 
            &              & 2 & 0.976 (0.02) & 0.965 (0.03) & 0.983 (0.01)   & 0.991 (0.00)  & 0.992 (0.00) & \bf 0.996 (0.00)  & 0.966     \\\cmidrule{3-10}  
            & \mr{2}{10k}  & 1 & 0.931 (0.09) & 0.979 (0.02) & 0.964 (0.03)   & 0.995 (0.01)  & \bf 0.998 (0.00) & 0.997 (0.01)  & \bf 0.998     \\                
            &              & 2 & 0.980 (0.04) & 0.998 (0.00) & \bf 0.999 (0.00) & 0.998 (0.00)& 0.998 (0.00) & 0.997 (0.01)      & 0.994     \\\cmidrule{3-10}   
            & \mr{2}{100k} & 1 & 0.909 (0.11) & 0.864 (0.12) & 0.849 (0.11)   & 0.995 (0.01)  & 0.997 (0.00) & 0.997 (0.00)      & \bf 1.000     \\ 
            &              & 2 & 0.976 (0.05) & 0.986 (0.02) & 0.980 (0.03)   & 0.999 (0.00)  & \bf 1.000 (0.00) & \bf 1.000 (0.00) & \bf 1.000     \\\bottomrule
\end{tabular}
}
\end{table*}
 
\begin{table*}[t]
\centering
\caption{Accuracy on Target SP Stringsets after 100 Epochs}
{\footnotesize
\label{tab:resultsSP}
\setlength\tabcolsep{4.5pt}
\begin{tabular}{ccc|ccc|ccc|c}
\toprule
\multicolumn{2}{c}{\mr{2}{Training}} & \mr{2}{Test} & 
\multicolumn{3}{c|}{LSTM}            & \multicolumn{3}{c|}{s-RNN}
                                     & \mr{2}{RPNI} \\[0.5ex] 

            &              &   & 10           & 30           & 100          & 10           & 30           & 100           &            \\\midrule
\mr{6}{SP2} & \mr{2}{1k}   & 1 & 0.847 (0.06) & 0.935 (0.07) & 0.952 (0.07) & 0.910 (0.05) & 0.999 (0.00)& 0.999 (0.00) & \bf 1.000 \\
            &              & 2 & 0.873 (0.10) & 0.951 (0.08) & 0.947 (0.08) & 0.976 (0.01) & \bf 1.000 (0.00) & \bf 1.000 (0.00) & \bf 1.000 \\\cmidrule{3-10}     
            & \mr{2}{10k}  & 1 & 0.734 (0.12) & 0.673 (0.04) & 0.720 (0.03) & 0.937 (0.13) & 0.960 (0.08) & 0.972 (0.06)  & \bf 1.000 \\                
            &              & 2 & 0.723 (0.12) & 0.656 (0.04) & 0.701 (0.03) & 0.934 (0.13) & 0.960 (0.08) & 0.972 (0.06)  & \bf 1.000 \\\cmidrule{3-10}   
            & \mr{2}{100k} & 1 & 0.680 (0.08) & 0.707 (0.10) & 0.732 (0.07) & 0.974 (0.07) & 0.974 (0.07) & 0.982 (0.04)  & \bf 1.000 \\  
            &              & 2 & 0.665 (0.09) & 0.697 (0.12) & 0.716 (0.08) & 0.977 (0.07) & 0.974 (0.08) & 0.986 (0.04)  & \bf 1.000 \\\midrule
\mr{6}{SP4} & \mr{2}{1k}   & 1 & 0.883 (0.06) & 0.885 (0.08) & 0.775 (0.05) & 0.890 (0.05) & 0.969 (0.02) & 0.988 (0.01)  & \bf 1.000 \\ 
            &              & 2 & 0.943 (0.04) & 0.840 (0.09) & 0.749 (0.06) & 0.885 (0.06) & 0.975 (0.01) & 0.985 (0.01)  & \bf 1.000 \\\cmidrule{3-10}       
            & \mr{2}{10k}  & 1 & 0.862 (0.13) & 0.880 (0.14) & 0.853 (0.08) & 0.696 (0.05) & 0.840 (0.15) & 0.903 (0.12)  & \bf 1.000 \\                       
            &              & 2 & 0.862 (0.14) & 0.877 (0.15) & 0.843 (0.08) & 0.686 (0.05) & 0.841 (0.16) & 0.900 (0.12)  & \bf 1.000 \\\cmidrule{3-10}       
            & \mr{2}{100k} & 1 & 0.842 (0.13) & 0.791 (0.14) & 0.720 (0.09) & 0.884 (0.14) & 0.828 (0.17) & 0.895 (0.12)  & \bf 1.000 \\ 
            &              & 2 & 0.831 (0.13) & 0.785 (0.13) & 0.716 (0.08) & 0.900 (0.15) & 0.827 (0.17) & 0.902 (0.13)  & \bf 1.000 \\\midrule
\mr{6}{SP8} & \mr{2}{1k}   & 1 & 0.844 (0.04) & 0.863 (0.05) & \bf 0.901 (0.01) & 0.871 (0.01) & 0.885 (0.02) & 0.878 (0.01)  & 0.817     \\ 
            &              & 2 & 0.699 (0.08) & 0.627 (0.05) & 0.692 (0.03) & \bf 0.719 (0.02) & 0.663 (0.06) & 0.668 (0.03)  & 0.587     \\\cmidrule{ 3-10}
            & \mr{2}{10k}  & 1 & 0.827 (0.15) & 0.798 (0.11) & 0.804 (0.04) & 0.818 (0.12) & 0.856 (0.10) & \bf 0.979 (0.02)  & 0.873     \\ 
            &              & 2 & 0.654 (0.11) & 0.672 (0.10) & 0.638 (0.05) & 0.566 (0.05) & 0.646 (0.05) & \bf 0.811 (0.08)  & 0.634     \\\cmidrule{3-10}  
            & \mr{2}{100k} & 1 & 0.880 (0.10) & 0.927 (0.08) & 0.904 (0.08) & 0.893 (0.14) & 0.978 (0.04) & 0.988 (0.01)  & \bf 1.000 \\ 
            &              & 2 & 0.760 (0.12) & 0.802 (0.13) & 0.739 (0.09) & 0.825 (0.15) & 0.909 (0.11) & 0.907 (0.09)  & \bf 1.000 \\\bottomrule
\end{tabular}
}
\end{table*}

The networks were run 10 times and we report the mean accuracy results on each test set, along with the standard deviation. 
RPNI, being deterministic, was only run once.
These results for the LSTMs and the s-RNNs and RPNI are given in Table~\ref{tab:resultsSL} for the SL targets and in Table~\ref{tab:resultsSP} for the SP targets.
In these tables, each row corresponds to one test set with one training set for one target stringset.
The mean value is given in each cell to three significant digits.
The standard deviation is given in parantheses to two significant digits.\footnote{This means of course that values close, but not equal to, one and zero may be shown as one and zero respectively. We confirm this happens in a few instances.}
Boldfaced numbers in each row indicate the best performance. 
Chance performance is 50\%. 

RPNI successfully output the target grammar in 10 of the 18 training regimens. These were the SL2 10k, SL2 100k, SL4 100k, SP8 100k, and all of the SP2 and SP4 training sets.\footnote{Even though the SL8 100k experiments shows RPNI with an accuracy of 1.000, this is due to the rounding error. The DFA RPNI output was not identical to the minimal target DFA.}

First we explain the results generally and then we make more specific comparisons. In all experiments the LSTMs and s-RNNs scored above chance indicating some learning took place. However, their mean results only outperformed RPNI in 9 of the 36 test sets. These are Test1 and Test2 in the SL4 1k, SL8 1k, SP8 1k and SP8 10k experiments, in addition to Test2 in the SL8 10k experiment. 

Also, the accuracies on Test1 and Test2 are nearly the same for the LSTMs and s-RNNs in almost all experiments except for the SP8 experiments.\footnote{The s-RNN performance is worse on Test2 than Test1 in the SL4 1k experiment.} 
On every SP8 experiment, the LSTM and s-RNN mean performance on Test1 is at least 10\% higher than the mean performance on Test2 (except for the s-RNNs on the SP8 100k experiments). 
Thus with few important exceptions, the generalizations acquired by the LSTMs and s-RNNs in training extended to longer words with a  similar degree of accuracy. 

However, in the SP8 experiments, the notable drop in accuracy in Test2 as compared to Test1 indicates that the networks failed to generalize to longer words. 
In the SP8 1k and 10k experiments, this failure may be excusable because the RPNI results are similar. 
More precisely, the failure of RPNI in these cases indicates that the training data itself was plausibly insufficient for accurate learning to take place. 
However, this failure of the networks in the SP8 100k experiment cannot be so excused because RPNI scored 100\% on both test sets indicating that the training data was sufficiently rich for proper inferences to occur.

We are also specifically interested in the LSTM performance in the SL versus SP experiments. 
With the exception of SL2 and SP2 experiments, there is generally a drop in accuracy for LSTMs on the SP experiments. 
This is most evident comparing the SP8 experiments with the SL8 experiments. 
In particular, LSTM mean accuracy on the SL4 and SL8 test sets were generally better than their mean accuracy on the SP4 and SP8 test sets. 
(The one exception is the comparison of the SL8 1k to the SP8 1k experiments.) 
This poorer performance in the SP experiments challenges the narrative in the deep learning literature that LSTMs solve the problem of learning long-term dependencies.

We are also interested in comparing the s-RNN performance between the SL and SP experiments. 
Like the LSTMS, the s-RNNs generally performed worse in the SP 4 and 8 experiments than the SL 4 and 8 experiments. 

Next we compare the performance of the LSTMs to the performance of the s-RNNs in the SP experiments. 
Given the narrative that the LSTM architecture addresses a known problem for s-RNNs, we expected that the s-RNN performance would be worse than the LSTM performance on the SP experiments. 
However this was not the case. 
In fact, the s-RNN results are about the same, and in some cases superior, perhaps most notably in the SP8 100k experiments.
Consequently, this result also challenges the deep learning narrative regarding long-term dependencies.

We conclude this section by mentioning an additional striking and unexpected result. 
The mean accuracy of the LSTMS and s-RNNs on the SL2 experiments, which are arguably the simplest patterns to learn, were among the worst results in all the experiments. 
While the accuracy in the SL2 1k experiments could be due to insufficient training data (as indicated by the RPNI result), this rationale is not available in the 10k and 100k experiments. 
Their performance here suggests that even the ability of recurrent networks to learn local dependencies in formal languages needs further study. 

\section{Discussion}
\label{sec:disc}

The results above indicate the the neural networks had the most difficulty in the SL2 and SP8 experiments.
Generally when learning fails, the possible culprits are either a deficiency in the data or some deficiency in the learning mechanism. The RPNI results exclude the former rationale for the SL2 10k, SL2 100k, and SP8 100k experiments. 

As for a deficiency in the neural networks, there are many possible modifications that would arguably improve outcomes. For instance, if the results are due the networks overfitting then the networks could be augmented with the dropout method or could be told to stop training before 100 epochs provided some condition is reached (``early stopping'').

We explored the possibility of early stopping and the results are shown in Tables~\ref{tab:resultsSLES} and~\ref{tab:resultsSPES}. These tables repeat the RPNI results for easier comparison.
\begin{table*}[t]
\centering
\caption{Accuracy on Target SL Stringsets Early Stopping}
{\footnotesize
\label{tab:resultsSLES}
\setlength\tabcolsep{4.5pt}
\begin{tabular}{ccc|ccc|ccc|c}
\toprule
\multicolumn{2}{c}{\mr{2}{Training}} & \mr{2}{Test} & 
\multicolumn{3}{c|}{LSTM}            & \multicolumn{3}{c|}{s-RNN}
									 & \mr{2}{RPNI} \\[0.5ex]

            &              &   & 10          & 30          & 100                & 10           & 30           & 100              &  \\\midrule
\mr{6}{SL2} & \mr{2}{1k}   & 1 & 0.818 (0.03) & 0.843 (0.05) & 0.923 (0.02)     & 0.848 (0.06) & 0.904 (0.03) & \bf 0.930 (0.03) & 0.855  \\ 
            &              & 2 & 0.780 (0.06) & 0.820 (0.07) & 0.905 (0.04)     & 0.871 (0.10) & 0.980 (0.02) & \bf 0.992 (0.01) & 0.844 \\\cmidrule{3-10} 
            & \mr{2}{10k}  & 1 & 0.925 (0.07) & 0.851 (0.04) & 0.875 (0.04)     & 0.936 (0.05) & 0.884 (0.07) & 0.729 (0.12)     & \bf 1.000\\ 
            &              & 2 & 0.919 (0.09) & 0.836 (0.06) & 0.835 (0.10)     & 0.964 (0.07) & 0.868 (0.11) & 0.753 (0.15)     & \bf 1.000\\\cmidrule{3-10} 
            & \mr{2}{100k} & 1 & 0.737 (0.14) & 0.711 (0.14) & 0.730 (0.03)     & 0.869 (0.15) & 0.767 (0.17) & 0.625 (0.01)     & \bf 1.000\\ 
            &              & 2 & 0.727 (0.15) & 0.698 (0.15) & 0.711 (0.04)     & 0.885 (0.17) & 0.766 (0.19) & 0.605 (0.01)     & \bf 1.000\\\midrule
\mr{6}{SL4} & \mr{2}{1k}   & 1 & 0.898 (0.01) & 0.939 (0.01) & 0.945 (0.01)     & 0.908 (0.01) & 0.945 (0.01) & \bf 0.958 (0.01) & 0.918\\ 
            &              & 2 & 0.829 (0.01) & 0.888 (0.01) & 0.887 (0.00)     & 0.840 (0.01) & 0.883 (0.01) & \bf 0.898 (0.01) & 0.813\\\cmidrule{3-10} 
            & \mr{2}{10k}  & 1 & 0.953 (0.05) & 0.956 (0.04) & \bf 0.997 (0.00) & 0.976 (0.03) & 0.982 (0.00) & 0.981 (0.00)     & 0.995\\
            &              & 2 & 0.934 (0.06) & 0.932 (0.05) & \bf 0.995 (0.01) & 0.973 (0.04) & 0.989 (0.00) & 0.990 (0.00)     & 0.978 \\\cmidrule{3-10}  
            & \mr{2}{100k} & 1 & 0.994 (0.00) & 0.973 (0.07) & \bf 1.000 (0.00) & 0.990 (0.00) & 0.990 (0.00) & 0.987 (0.00)     & \bf 1.000\\ 
            &              & 2 & 0.996 (0.00) & 0.975 (0.07) & \bf 1.000 (0.00) & 0.996 (0.00) & 0.996 (0.00) & 0.995 (0.00)     & \bf 1.000\\\midrule
\mr{6}{SL8} & \mr{2}{1k}   & 1 & 0.966 (0.02) & 0.983 (0.01) & \bf 0.995 (0.00) & 0.962 (0.02) & 0.969 (0.01) & 0.971 (0.01)     & 0.991\\ 
            &              & 2 & 0.971 (0.01) & 0.980 (0.02) & \bf 0.994 (0.00) & 0.969 (0.02) & 0.974 (0.01) & 0.977 (0.01)     & 0.966\\\cmidrule{3-10}  
            & \mr{2}{10k}  & 1 & 0.990 (0.01) & 0.996 (0.00) & \bf 0.999 (0.00) & 0.998 (0.00) & \bf 0.999 (0.00) & \bf 0.999 (0.00) & 0.998 \\         
            &              & 2 & 0.994 (0.00) & 0.995 (0.00) & \bf 0.998 (0.00) & 0.997 (0.00) & \bf 0.998 (0.00) & \bf 0.998 (0.00) & 0.994\\\cmidrule{3-10}   
            & \mr{2}{100k} & 1 & 0.993 (0.01) & 0.998 (0.00) & \bf 1.000 (0.00) & 0.999 (0.00) & \bf 1.000 (0.00) & \bf 1.000 (0.00) & \bf 1.000\\ 
            &              & 2 & 0.994 (0.01) & 0.999 (0.00) & \bf 1.000 (0.00) & 0.999 (0.00) & \bf 1.000 (0.00) & \bf 1.000 (0.00) & \bf 1.000\\\bottomrule
\end{tabular}
}
\end{table*}

Training for the networks were halted 
at the epoch which maximized the accuracy of the networks on the validation set. 
The validation set was a held-out fraction (10\%) of the training set that was randomly selected from the training set. Because the training set contained duplicates, it is not the case that every test item in the validation set is novel, as is the case with Test1 and Test2 items.

\begin{table*}[t]
\centering
\caption{Accuracy on Target SP Stringsets Early Stopping}
{\footnotesize
\label{tab:resultsSPES}
\setlength\tabcolsep{4.5pt}
\begin{tabular}{ccc|ccc|ccc|c}
\toprule
\multicolumn{2}{c}{\mr{2}{Training}} & \mr{2}{Test} & 
\multicolumn{3}{c|}{LSTM}            & \multicolumn{3}{c|}{s-RNN}
                                     & \mr{2}{RPNI} \\[0.5ex]

            &              &   & 10           & 30           & 100           & 10           & 30           & 100                &          \\\midrule
\mr{6}{SP2} & \mr{2}{1k}   & 1 & 0.871 (0.04) & 0.954 (0.05) & 0.992 (0.00) & 0.910 (0.05) & 0.994 (0.01) & 0.992 (0.01)   & \bf 1.000  \\
            &              & 2 & 0.960 (0.03) & 0.989 (0.02) & 0.998 (0.00) & 0.976 (0.01) & 0.998 (0.01) & \bf 1.000 (0.00)   & \bf 1.000 \\\cmidrule{3-10}     
            & \mr{2}{10k}  & 1 & 0.890 (0.07) & 0.941 (0.04) & 0.977 (0.02)  & 0.995 (0.01) & 0.981 (0.05) & 0.999 (0.00)  & \bf 1.000   \\                
            &              & 2 & 0.979 (0.02) & 0.990 (0.01) & 0.994 (0.01)  & \bf 1.000 (0.00) & 0.984 (0.05) & \bf 1.000 (0.00) & \bf 1.000  \\\cmidrule{3-10}   
            & \mr{2}{100k} & 1 & 0.833 (0.14) & 0.819 (0.12) & 0.890 (0.08)  & 0.997 (0.01) & 0.999 (0.00) & 0.997 (0.00) & \bf 1.000   \\  
            &              & 2 & 0.838 (0.16) & 0.805 (0.13) & 0.872 (0.09)  & \bf 1.000 (0.00) & \bf 1.000 (0.00) & \bf 1.000 (0.00) & \bf 1.000  \\\midrule
\mr{6}{SP4} & \mr{2}{1k}   & 1 & 0.881 (0.06) & 0.946 (0.04) & 0.963 (0.03)  & 0.887 (0.05) & 0.966 (0.02) & 0.979 (0.01)   & \bf 1.000   \\ 
            &              & 2 & 0.950 (0.03) & 0.960 (0.03) & 0.983 (0.01)  & 0.883 (0.05) & 0.975 (0.01) & 0.979 (0.01)   & \bf 1.000  \\\cmidrule{3-10}       
            & \mr{2}{10k}  & 1 & 0.899 (0.11) & 0.958 (0.07) & 0.991 (0.01)  & 0.935 (0.08) & 0.968 (0.04) & 0.999 (0.00)  & \bf 1.000   \\                       
            &              & 2 & 0.926 (0.09) & 0.971 (0.05) & 0.991 (0.01)  & 0.954 (0.07) & 0.984 (0.02) & \bf 1.000 (0.00) & \bf 1.000  \\\cmidrule{3-10}       
            & \mr{2}{100k} & 1 & 0.943 (0.08) & 0.940 (0.08) & 0.920 (0.06)  & 0.942 (0.09) & 0.958 (0.09) & 0.973 (0.07)   & \bf 1.000\\ 
            &              & 2 & 0.928 (0.08) & 0.930 (0.09) & 0.911 (0.07)  & 0.951 (0.09) & 0.962 (0.08) & 0.974 (0.08)   & \bf 1.000 \\\midrule
\mr{6}{SP8} & \mr{2}{1k}   & 1 & 0.884 (0.02) & 0.884 (0.02) & \bf 0.903 (0.02)  & 0.861 (0.01) & 0.878 (0.02) & 0.857 (0.02)   & 0.817   \\ 
            &              & 2 & \bf 0.733 (0.03) & 0.643 (0.06) & 0.688 (0.04)  & 0.730 (0.01) & 0.681 (0.06) & 0.625 (0.04)   & 0.587  \\\cmidrule{ 3-10}
            & \mr{2}{10k}  & 1 & 0.934 (0.05) & 0.921 (0.05) & 0.959 (0.03)  & 0.908 (0.02) & 0.952 (0.03) & \bf 0.991 (0.00)  & 0.873  \\ 
            &              & 2 & 0.637 (0.08) & 0.659 (0.10) & 0.704 (0.11)  & 0.600 (0.08) & 0.640 (0.10) & \bf 0.837 (0.05)   & 0.634 \\\cmidrule{3-10}  
            & \mr{2}{100k} & 1 & 0.977 (0.04) & 0.975 (0.04) & 0.980 (0.02)  & 0.964 (0.05) & 0.990 (0.03) & \bf 1.000 (0.00) & \bf 1.000 \\ 
            &              & 2 & 0.881 (0.11) & 0.865 (0.13) & 0.864 (0.08)  & 0.890 (0.08) & 0.942 (0.09) & 0.984 (0.03)   & \bf 1.000 \\\bottomrule
\end{tabular}
}
\end{table*}

These experiments show that the SL2 results do improve, but not to the level of the other experimental conditions. 
There is also improvement in the SP8 experiments though the difference between Test1 and Test2 results remains large. We also experimented with the dropout method (not shown). Like early stopping, this improved the results, but not the observed trends described above.	

Of course there are other ways to improve the "learning power" of recurrent neural networks. 
The vector sizes can be increased, multiple layers can be included in the recurrent components, Kalman filters could be used, and so on. 
So this is one avenue of future research. 

However, these methods go hand-in-hand with varying the complexity of the target languages at different levels of abstraction, such as adding more forbidden strings to the grammars, increasing the size of the alphabet, increasing $k$,  or moving up the subregular hierarchy to more complex classes.
The goal here is not one-upmanship, but to instead get a better understanding of how properties of RNNs relate to properties of formal language classes, like the well-understood subregular ones presented in section~\ref{sec:subreg}.

As such, these experiments showed something very clearly. They showed naive LSTMs have difficulty with learning a formal language defined by a forbidden subsequence of length 8 but little to no difficulty with learning a formal language defined by forbidden substrings of length 8.
The difference between substring and subsequence---which reduces logically to the question of whether order is represented in strings with the successor or precedence relation \cite{Rogers-HeinzEtAl-2013-CSC}---is thus significant for naive LSTMs. 

The successor/precedence difference is also challenging for s-RNNs, but they are not any more challenging for s-RNNs than they are for LSTMs. In fact, s-RNNs outperform LSTMs on many of the SP experiments.

From these results, it is hard to see how LSTMs are a solution to learning long-term dependencies, which are problematic for s-RNNs. 

\citet{Rodriguez2001} cautions against underestimating s-RNNs. In this article he argues he shows ``a range of language tasks in which an [s-RNN] develops solutions that not only count but also copy and store counting information. 
In one case, the network stores information like an explicit storage mechanism. 
In other cases, the network stores information more indirectly\ldots'' In short, despite the well-known exploding and vanishing gradient problems, s-RNNs can learn to store information long-term in some circumstances.

So what then explains the unexpected performance of s-RNNs? 
One possibility is the Adam optimization method \citep{KingmaB14} which the s-RNNs in these experiments used. 
This optimization technique is relatively recent and replaces the method of stochastic gradient descent (SGD), which was used in the decades prior to Adam's introduction. 
The narrative regarding long-term dependencies surrounding LSTMs and s-RNNs was developed in the context of SGD.
So one hypothesis is that the narrative is conditioned on the use of SGD as the optimization method, but that once Adam is used in its place, the capacity of s-RNNs to learn long-term dependencies is much improved. In a sense, maybe Adam itself goes some distance in resolving the vanishing and exploding gradient problems.

We have begun to test this hypothesis by running the experiments with SGD instead of Adam. 
While we are not yet able to provide a full report, the result of a single experiment with s-RNNs using SGD as the optimization method shows a serious decline its performance as shown in Table~\ref{tab:SRNN-SGD}
\begin{table}[ht]
\centering
\caption{s-RNN with SGD Accuracy on Target Stringsets after 100 epochs}
{\scriptsize
\label{tab:SRNN-SGD}
\begin{tabular}{ccccc}
\toprule
    &          & \multicolumn{3}{c}{sRNN with SGD}                        \\ \cmidrule{3-5}
    & Training & v10           & v30            & v100              \\ \cmidrule{3-5}
    & Regimen  & Test1  Test2  & Test1  Test2   & Test1  Test2      \\ \midrule
    & 1k       & 0.7641 0.7228 & 0.8015 0.7336  & 0.7362 0.6898    \\ \cmidrule{2-5}
SL2 & 10k      & 0.7194 0.6983 & 0.7752 0.7169  & 0.9745 0.9981     \\ \cmidrule{2-5}
    & 100k     & 0.8034 0.7417 & 0.9713 0.9984  & 0.9482 0.9968     \\ \midrule\midrule
    & 1k       & 0.9304 0.8683 & 0.9072 0.8359  & 0.8789 0.8134\\ \cmidrule{2-5}
SL4 & 10k      & 0.8234 0.7870 & 0.8347 0.7977  & 0.8588 0.8187 \\ \cmidrule{2-5}
    & 100k     & 0.4738 0.5005 & 0.8957 0.8492  & 0.5567 0.5641 \\ \midrule\midrule
    & 1k       & 0.8191 0.8265 & 0.8402 0.8564  & 0.8065 0.8064 \\ \cmidrule{2-5}
SL8 & 10k      & 0.5076 0.4951 & 0.5076 0.4951  & 0.6441 0.6409 \\ \cmidrule{2-5}   
    & 100k     & 0.7830 0.7789 & 0.7830 0.7789  & 0.8079 0.8031 \\ \midrule\midrule
    & 1k       & 0.7346 0.7331 & 0.8739 0.9174  & 0.8370 0.9277 \\ \cmidrule{2-5}    
SP2 & 10k      & 0.8371 0.8449 & 0.8400 0.8264  & 0.9998 1.0000 \\ \cmidrule{2-5}   
    & 100k     & 0.6071 0.5902 & 0.5696 0.5645  & 1.0000 1.0000 \\ \midrule\midrule 
    & 1k       & 0.7625 0.7361 & 0.5744 0.5306  & 0.6668 0.5991\\ \cmidrule{2-5}   
SP4 & 10k      & 0.7270 0.6808 & 0.6167 0.6331  & 0.7270 0.6398 \\ \cmidrule{2-5}   
   & 100k      & 0.7038 0.6287 & 0.5626 0.5582  & 0.6643 0.5845 \\ \midrule\midrule 
  & 1k         & 0.8160 0.7702 & 0.1984 0.2882  & 0.8393 0.7372 \\ \cmidrule{2-5}   
SP8 & 10k      & 0.6520 0.6201 & 0.8364 0.6677  & 0.4504 0.5204 \\ \cmidrule{2-5}   
    & 100k     & 0.8251 0.7505 & 0.8117 0.7645  & 0.4409 0.5027 \\ \bottomrule
\end{tabular}
}
\end{table}
Obviously, this is an area of current and future research.

Another possibility is highlighted by the fact that there are different \emph{kinds} of long-term dependencies. It may be that LSTMs outperform s-RNNs on long-term dependencies unlike the Strictly Piecewise ones tested here. As mentioned, the subregular hierarchies (Figure~\ref{fig:subreg}) are particularly good at distinguishing different types of long-term dependencies. One contribution of this paper is the more fine-grained classification of long-term dependencies that formal language theory and the subregular language classes provide researchers.

Finally, we return to the discussion of interpreting the performance of the neural networks in light of the grammatical inference algorithm RPNI. 
We argued it was useful to use RPNI to evaluate the quality of the data.
However, this argument has one potential flaw. 
RPNI can be said to measure the sufficiency of the data, at least for learning mechanisms which are targeting \emph{regular} stringsets. 
If the learning mechanisms are capable of learning nonregular languages then they may need \emph{more} data to learn some regular languages (in order to distinguish them from some context-free language for example). 

RPNI in a sense ``knows'' that it is learning regular stringsets because it only ever builds finite-state acceptors.
There is evidence that RNNs, on the other hand, can learn some non-regular stringsets \citep{Rodriguez2001, Schmidhuber2002}. 
Furthermore, it is not known how to build this kind of a priori knowledge into neural networks. 
In other words, the use of RPNI to understand the quality of the training data is suggestive but not probative.
Nonetheless, in the absence of theoretical learning results of RNNs on nonregular formal languages, using grammatical inference  algorithms which provably learn large classes of languages (like RPNI) may be the best anyone can do.

\section{Conclusion}
\label{sec:c}


In this paper, we developed controlled experiments using formal languages for investigating the ability of RNNs to learn long-term dependencies. The results are difficult to understand in light of the dominant narrative in 
the deep learning literature regarding the efficacy of LSTMs over s-RNNs in this learning task.

More generally, the experiments presented here help show how formal language theory can reveal the advantages and disadvantages of various RNN models more clearly than testing with real-world datasets.
The primary reason is that we control the nature of the target patterns and their complexity. 
This was illustrated here with the comparison of SL and SP languages which encode local and long-term dependencies, respectively. 
From a logical perspective, the only difference between the SL and SP classes is the way in which order is represented in strings: SL classes use the successor relation and SP classes use the precedence relation. 

Controlled experiments can also include carefully designed test sets. Here, through Test1 and Test2, we could control the test data to better understand how the RNNs generalize to words longer than the ones found in training.

We argued these controlled experiments show there is still some distance to go to understand how RNNs, including LSTMs, represent and learn long-term dependencies in sequential patterns. 
We tentatively hypothesized that the optimization technique Adam may alleviate some difficulties  that s-RNNs may have with long-term dependencies. If correct, this means the source of the problem was not the s-RNNs per se, but SGD. 

Finally, we believe that more controlled experiments of the sort presented here with more complex formal languages and less naive LSTMs and other kinds of RNNs, in conjunction with learning algorithms from grammatical inference, are critical to better understanding the capacity of neural networks to represent and learn long-term dependencies.

\section*{Acknowledgements}
The authors acknowledge the support of the Japan Society for the
Promotion of Science (JSPS) KAKENHI grant 26730123 to CS and the support of the National Institute of Health (NIH) grant R01HD087133-01 to JH.

We are grateful to the feedback of the audiences of the LAML conference, the LearnAut workshop and the IACS seminar. We also thank Jim Rogers for assistance in determining the complexity of the embedded Reber grammar.

\bibliographystyle{acl_natbib}
\bibliography{main}

\begin{thebibliography}{}
\expandafter\ifx\csname natexlab\endcsname\relax\def\natexlab#1{#1}\fi

\bibitem[{Akram et~al.(2010)Akram, de~la Higuera, Xiao, and Eckert}]{Akram2010}
Hasan~Ibne Akram, Colin de~la Higuera, Huang Xiao, and Claudia Eckert. 2010.
\newblock Grammatical inference algorithms in matlab.
\newblock In Jos{\'e}~M. Sempere and Pedro Garc{\'i}a, editors, {\em
  Proceedings of the 10th International Colloquium of Grammatical Inference:
  Theoretical Results and Applications\/}, Springer Berlin Heidelberg, Berlin,
  Heidelberg, pages 262--266.

\bibitem[{Bengio et~al.(1994)Bengio, Simard, and Frasconi}]{BENGIO1994}
Y.~Bengio, P.~Simard, and P.~Frasconi. 1994.
\newblock Learning long-term dependencies with gradient descent is difficult.
\newblock {\em IEEE Transactions on Neural Networks\/} 5(2):157--166.

\bibitem[{B{\"u}chi(1960)}]{Buchi1960}
J.~Richard B{\"u}chi. 1960.
\newblock Weak second-order arithmetic and finite automata.
\newblock {\em Mathematical Logic Quarterly\/} 6(1-6):66--92.

\bibitem[{Casey(1996)}]{Casey1996}
Mike Casey. 1996.
\newblock The dynamics of discrete-time computation with application to
  recurrent neural networks and finite state machine extraction.
\newblock {\em Neural computation\/} 8(6):1135--1178.

\bibitem[{Chalup and Blair(2003)}]{Chalup2003955}
Stephan~K. Chalup and Alan~D. Blair. 2003.
\newblock Incremental training of first order recurrent neural networks to
  predict a context-sensitive language.
\newblock {\em Neural Networks\/} 16(7):955--972.

\bibitem[{Chomsky(1956)}]{chomsky56}
Noam Chomsky. 1956.
\newblock Three models for the description of language.
\newblock {\em IRE Transactions on Information Theory\/} page 113–124.
\newblock IT-2.

\bibitem[{Chomsky(1957)}]{chomsky57}
Noam Chomsky. 1957.
\newblock {\em Syntactic Structures\/}.
\newblock Mouton \& Co., Printers, The Hague.

\bibitem[{Chomsky(1965)}]{chomsky65}
Noam Chomsky. 1965.
\newblock {\em Aspects of the theory of syntax\/}.
\newblock Cambridge, MA: MIT Press.

\bibitem[{de~la Higuera(1997)}]{Higuera1997}
Colin de~la Higuera. 1997.
\newblock Characteristic sets for polynomial grammatical inference.
\newblock {\em Machine Learning\/} 27(2):125--138.

\bibitem[{de~la Higuera(2010)}]{Higuera2010}
Colin de~la Higuera. 2010.
\newblock {\em Grammatical Inference: Learning Automata and Grammars\/}.
\newblock Cambridge University Press.

\bibitem[{Elman(1990)}]{ELMAN1990179}
Jeffrey~L. Elman. 1990.
\newblock Finding structure in time.
\newblock {\em Cognitive Science\/} 14(2):179--211.

\bibitem[{Enderton(2001)}]{Enderton2001}
Herbert~B. Enderton. 2001.
\newblock {\em A Mathematical Introduction to Logic\/}.
\newblock Academic Press, 2nd edition.

\bibitem[{Eyraud et~al.(2016)Eyraud, Heinz, and
  Yoshinaka}]{Eyraud-HeinzEtAL-2016-EILLP}
R{\'e}mi Eyraud, Jeffrey Heinz, and Ryo Yoshinaka. 2016.
\newblock Efficiency in the identification in the limit learning paradigm.
\newblock In Jeffrey Heinz and Jos{\'e} Sempere, editors, {\em Topics in
  Grammatical Inference\/}, Springer-Verlag Berlin Heidelberg, pages 25--46.

\bibitem[{Garc{\'i}a and Ruiz(2004)}]{Garca2004LearningKA}
Pedro Garc{\'i}a and Jos{\'e} Ruiz. 2004.
\newblock Learning k-testable and k-piecewise testable languages from positive
  data.
\newblock {\em Grammars\/} 7:125--140.

\bibitem[{Garcia et~al.(1990)Garcia, Vidal, and Oncina}]{GarciaEtAl1990}
Pedro Garcia, Enrique Vidal, and Jos{\'e} Oncina. 1990.
\newblock Learning locally testable languages in the strict sense.
\newblock In {\em Proceedings of the Workshop on Algorithmic Learning
  Theory\/}. pages 325--338.

\bibitem[{Giles et~al.(1992)Giles, Miller, Chen, Chen, Sun, and
  Lee}]{Giles1992}
C~L Giles, C~B Miller, D~Chen, H~H Chen, G~Z Sun, and Y~C Lee. 1992.
\newblock {Learning and Extracting Finite State Automata with 2nd-Order
  Recurrent Neural Networks}.
\newblock {\em Neural Computation\/} 4(3):393--405.

\bibitem[{Goldberg(2017)}]{Goldberg2017}
Yoav Goldberg. 2017.
\newblock {\em Neural Network Methods for Natural Language Processing\/}.
\newblock Morgan and Claypool Publishers.

\bibitem[{Goodfellow et~al.(2016)Goodfellow, Bengio, and
  Courville}]{GoodfellowBengioCourville2016}
Ian Goodfellow, Yoshua Bengio, and Aaron Courville. 2016.
\newblock {\em Deep Learning\/}.
\newblock The MIT Press.

\bibitem[{Greff et~al.(2015)Greff, Srivastava, Koutn{\'{\i}}k, Steunebrink, and
  Schmidhuber}]{GreffSKSS15}
Klaus Greff, Rupesh~Kumar Srivastava, Jan Koutn{\'{\i}}k, Bas~R. Steunebrink,
  and J{\"{u}}rgen Schmidhuber. 2015.
\newblock {LSTM:} {A} search space odyssey.
\newblock {\em CoRR\/} abs/1503.04069.

\bibitem[{Heinz(2010{\natexlab{a}})}]{Heinz-2010-LLP}
Jeffrey Heinz. 2010{\natexlab{a}}.
\newblock Learning long-distance phonotactics.
\newblock {\em Linguistic Inquiry\/} 41(4):623--661.

\bibitem[{Heinz(2010{\natexlab{b}})}]{Heinz-2010-SEL}
Jeffrey Heinz. 2010{\natexlab{b}}.
\newblock String extension learning.
\newblock In {\em Proceedings of the 48th Annual Meeting of the Association for
  Computational Linguistics\/}. Association for Computational Linguistics,
  Uppsala, Sweden, pages 897--906.

\bibitem[{Heinz et~al.(2015)Heinz, de~la Higuera, and van
  Zaanen}]{Heinz-delaHiguera-vanZaanen-GICL}
Jeffrey Heinz, Colin de~la Higuera, and Menno van Zaanen. 2015.
\newblock {\em Grammatical Inference for Computational Linguistics\/}.
\newblock Synthesis Lectures on Human Language Technologies. Morgan and
  Claypool.

\bibitem[{Heinz et~al.(2012)Heinz, Kasprzik, and
  K{\"{o}}tzing}]{Heinz-KasprzikEtAl-2012-LLHS}
Jeffrey Heinz, Anna Kasprzik, and Timo K{\"{o}}tzing. 2012.
\newblock Learning with lattice-structured hypothesis spaces.
\newblock {\em Theoretical Computer Science\/} 457:111--127.

\bibitem[{Hochreiter and Schmidhuber(1997)}]{lstm1997}
S~Hochreiter and J~Schmidhuber. 1997.
\newblock Long short-term memory.
\newblock {\em Neural Computation\/} 9(8):1735--1780.

\bibitem[{Hulden(2009)}]{hulden2009-foma}
Mans Hulden. 2009.
\newblock Foma: a finite-state compiler and library.
\newblock In {\em Proceedings of the 12th Conference of the European Chapter of
  the Association for Computational Linguistics\/}. Association for
  Computational Linguistics, pages 29--32.

\bibitem[{Jurafsky and Martin(2008)}]{JM2008}
Daniel Jurafsky and James Martin. 2008.
\newblock {\em Speech and Language Processing: An Introduction to Natural
  Language Processing, Speech Recognition, and Computational Linguistics\/}.
\newblock Prentice-Hall, Upper Saddle River, NJ, 2nd edition.

\bibitem[{Kingma and Ba(2014)}]{KingmaB14}
Diederik~P. Kingma and Jimmy Ba. 2014.
\newblock Adam: {A} method for stochastic optimization.
\newblock {\em CoRR\/} abs/1412.6980.

\bibitem[{LeCun et~al.(2015)LeCun, Bengio, and Hinton}]{deep}
Yann LeCun, Yoshua Bengio, and Geoffrey Hinton. 2015.
\newblock Deep learning.
\newblock {\em Nature\/} 521(7553):436--444.

\bibitem[{McNaughton and Papert(1971)}]{McNaughtonPapert1971}
Robert McNaughton and Seymour Papert. 1971.
\newblock {\em Counter-Free Automata\/}.
\newblock MIT Press.

\bibitem[{Oncina and Garcia(1992)}]{OncinaGarcia1992-RPNI}
Jose Oncina and Pedro Garcia. 1992.
\newblock Identifying regular languages in polynomial time.
\newblock In {\em Advances In Structural And Syntactic Pattern Recognition,
  Volume 5 Of Series In Machine Perception And Artificial Intelligence\/}.
  World Scientific, pages 99--108.

\bibitem[{Pollack(1991)}]{Pollack1991}
Jordan~B. Pollack. 1991.
\newblock {The Induction of Dynamical Recognizers}.
\newblock {\em Machine Learning\/} 7(2):227--252.

\bibitem[{Pérez-Ortiz et~al.(2003)Pérez-Ortiz, Gers, Eck, and
  Schmidhuber}]{PérezOrtiz2003241}
Juan~Antonio Pérez-Ortiz, Felix~A. Gers, Douglas Eck, and Jürgen Schmidhuber.
  2003.
\newblock Kalman filters improve \{LSTM\} network performance in problems
  unsolvable by traditional recurrent nets.
\newblock {\em Neural Networks\/} 16(2):241 -- 250.

\bibitem[{Reber(1967)}]{REBER1967}
Arthur~S. Reber. 1967.
\newblock Implicit learning of artificial grammars.
\newblock {\em Journal of Verbal Learning and Verbal Behavior\/} 6(6):855--863.

\bibitem[{Rodriguez(2001)}]{Rodriguez2001}
Paul Rodriguez. 2001.
\newblock Simple recurrent networks learn context-free and context-sensitive
  languages by counting.
\newblock {\em Neural Computation\/} 13(9):2093--2118.

\bibitem[{Rogers et~al.(2010)Rogers, Heinz, Bailey, Edlefsen, Visscher,
  Wellcome, and Wibel}]{Rogers-HeinzEtAl-2010-LPTSS}
James Rogers, Jeffrey Heinz, Gil Bailey, Matt Edlefsen, Molly Visscher, David
  Wellcome, and Sean Wibel. 2010.
\newblock On languages piecewise testable in the strict sense.
\newblock In Christian Ebert, Gerhard J{\"a}ger, and Jens Michaelis, editors,
  {\em The Mathematics of Language\/}. Springer, volume 6149 of {\em Lecture
  Notes in Artifical Intelligence\/}, pages 255--265.

\bibitem[{Rogers et~al.(2013)Rogers, Heinz, Fero, Hurst, Lambert, and
  Wibel}]{Rogers-HeinzEtAl-2013-CSC}
James Rogers, Jeffrey Heinz, Margaret Fero, Jeremy Hurst, Dakotah Lambert, and
  Sean Wibel. 2013.
\newblock Cognitive and sub-regular complexity.
\newblock In Glyn Morrill and Mark-Jan Nederhof, editors, {\em Formal
  Grammar\/}. Springer, volume 8036 of {\em Lecture Notes in Computer
  Science\/}, pages 90--108.

\bibitem[{Rogers and Pullum(2011)}]{RogersPullum2011}
James Rogers and Geoffrey Pullum. 2011.
\newblock Aural pattern recognition experiments and the subregular hierarchy.
\newblock {\em Journal of Logic, Language and Information\/} 20:329--342.

\bibitem[{Schmidhuber(2015)}]{Schmidhuber2015}
J.~Schmidhuber. 2015.
\newblock Deep learning in neural networks: An overview.
\newblock {\em Neural Networks\/} 61:85--117.

\bibitem[{Schmidhuber et~al.(2002)Schmidhuber, Gers, and Eck}]{Schmidhuber2002}
J.~Schmidhuber, F.~Gers, and D.~Eck. 2002.
\newblock Learning nonregular languages: A comparison of simple recurrent
  networks and lstm.
\newblock {\em Neural Computation\/} 14:2039--2041.

\bibitem[{Simon(1975)}]{Simon1975}
Imre Simon. 1975.
\newblock Piecewise testable events.
\newblock In {\em Automata Theory and Formal Languages 2nd GI Conference
  Kaiserslautern, May 20--23, 1975\/}. Springer, pages 214--222.

\bibitem[{{Smith, A.W.}(1989)}]{SmithA.W.1989}
Zipser~D. {Smith, A.W.} 1989.
\newblock {Encoding sequential structure: experience with the real-time
  recurrent learning algorithm}.
\newblock {\em Neural Networks\/} pages 0--4.

\bibitem[{Thomas(1982)}]{Thomas1982}
Wolfgang Thomas. 1982.
\newblock Classifying regular events in symbolic logic.
\newblock {\em Journal of Computer and Systems Sciences\/} 25:370--376.

\bibitem[{Tomita(1982)}]{Tomita1982}
Masaru Tomita. 1982.
\newblock {Learning of construction of finite automata from examples using
  hill-climbing.}
\newblock {\em Proc. Fourth Int. Cog. Sci. Conf.\/} pages 105 -- 108.

\bibitem[{Watrous and Kuhn(1992)}]{Watrous1992}
Raymond~L Watrous and G~M Kuhn. 1992.
\newblock {Induction of Finite-State Automata Using Second-Order Recurrent
  Networks}.
\newblock {\em Advances in Neural Information Processing Systems\/}
  (10):309--316.

\end{thebibliography}
\end{document}